\documentclass{article}

\usepackage{spconf,amsmath,graphicx}

\usepackage[utf8]{inputenc} %
\usepackage[T1]{fontenc}    %
\usepackage[hidelinks]{hyperref}       %
\usepackage{url}            %
\usepackage{booktabs}       %
\usepackage{amsfonts}       %
\usepackage{nicefrac}       %
\usepackage{microtype}      %
\usepackage{xcolor}         %
\usepackage{graphicx,epstopdf}
\usepackage{adjustbox}
\usepackage{multirow}
\usepackage{csquotes}
\usepackage{wrapfig}
\usepackage{makecell}  %

\usepackage{enumitem}
\usepackage{subcaption}
\usepackage{comment}

\renewcommand{\paragraph}[1]{\vspace{0.5em}\noindent\textbf{#1}\hspace{0.5mm}}

\usepackage{mathtools}

\usepackage{xspace}
\makeatletter
\DeclareRobustCommand\onedot{\futurelet\@let@token\@onedot}
\def\@onedot{\ifx\@let@token.\else.\null\fi\xspace}

\def\eg{\emph{e.g}\onedot} 
\def\ie{\emph{i.e}\onedot} 
 
\def\etc{\emph{etc}\onedot}

\makeatother

\usepackage{sistyle}
\SIthousandsep{,}

\usepackage{upgreek}

\def\mN{\mathcal{N}}

\def\mS{\mathcal{S}}

\def\1n{\mathbf{1}_n}
\def\0{\mathbf{0}}
\def\1{\mathbf{1}}

\def\R{{\mathbb R}}

\def\c{{\bf c}}

\def\x{{\bf x}}

\def\z{{\bf z}}

\def\bmu{\mbox{\boldmath{$\mu$}}}

\newcommand{\cX}{\mathcal{X}}

\usepackage{physics}

\usepackage{array}
\newcolumntype{C}[1]{>{\centering\arraybackslash}m{#1}}  %

\newcommand{\secsym}{Sec\onedot}
\newcommand{\figsym}{Fig\onedot}

\newcommand{\secref}[1]{\secsym~\ref{#1}}
\newcommand{\figref}[1]{\figsym~\ref{#1}}

\newcommand{\past}{{1..{t{-}1}}}  %

\usepackage{placeins}
\usepackage{xr}

\newcommand{\WER}[3]{{$\mbox{CER}^{\mbox{#1}}_{\, \mbox{#2} {\rightarrow} \mbox{#3}}$}\xspace}
\newcommand{\WERtestRR}{\WER{}{r}{r}}
\newcommand{\WERtestRS}{\WER{}{r}{s}}
\newcommand{\WERtestSR}{\WER{}{s}{r}}
\newcommand{\WERtestSS}{\WER{}{s}{s}}

\newcommand{\WERtestBR}{\WER{}{b}{r}}
\newcommand{\WERtestXY}{\WER{}{t}{v}}

\newcommand{\dx}{{\Delta x}}
\newcommand{\dy}{{\Delta y}}

\title{Data Incubation --- Synthesizing Missing Data for \\Handwriting Recognition }

\name{
	Jen-Hao Rick Chang, 
	Martin Bresler, 
	Youssouf Chherawala, 
	Adrien Delaye,
}
\secondlinename{
	Thomas Deselaers,
	Ryan Dixon,
	Oncel Tuzel	
}
\address{Apple}

\begin{document}
\maketitle

\begin{abstract}

In this paper, we demonstrate how a generative model can be used to build a better recognizer through the control of content and style.
We are building an online handwriting recognizer from a modest amount of training samples.
By training our controllable handwriting synthesizer on \emph{the same} data, we can synthesize handwriting with previously underrepresented content (\eg, URLs and email addresses) and style (\eg, cursive and slanted). 
Moreover, we propose a framework to analyze a recognizer that is trained with a mixture of real and synthetic training data. %
We use the framework to optimize data synthesis and demonstrate significant improvement on handwriting recognition over a model trained on real data only. 
Overall, we achieve a 66\% reduction in Character Error Rate.
\end{abstract}

\section{Introduction}
\label{sec: introduction}

\begin{figure}[th!]
	\centering
	\includegraphics[width=.8\linewidth]{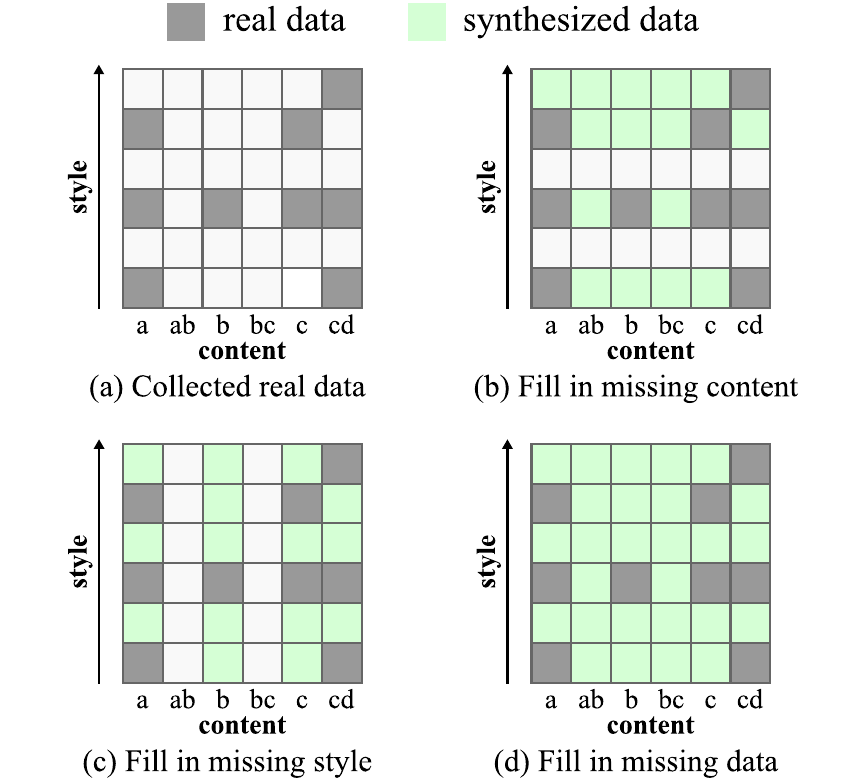}
	\caption{
	    Data incubation starts with (a) collected data sparsely covering the content and the style space.
	    We then use a controllable generative model that can (b) synthesize samples with unseen content, and (c) mix new style from existing training samples. 
	    (d) Finally, we synthesize new samples that close gaps in content and style, leading to a better coverage of the real data distribution for handwriting recognition. 
}	    
	\label{figure:illustration}
\end{figure}

Data is the most valuable asset for machine learning applications, but are we using it efficiently? 
The standard approach is relatively simple: (i) collect and annotate data, and (ii) train a recognizer using the data.
While this approach has achieved success in many standard machine learning tasks, such as image classification~\cite{tan2019efficientnet}, we will show that it is sub-optimal for handwriting recognition.

Handwriting data can be described 
(a) by its content (\ie, the text that was written, a sequence of characters)
and (b) by its style (\ie, all other information about a handwriting sample except content {\textemdash} printed or cursive, neat or messy, round or square corners, tight or loose spacing, \etc).
The content space is combinatorial and some handwriting styles are rare compared to others. 
Therefore, it is difficult to collect sufficient data to cover the entire content and style space well.

In this paper, we propose a method we refer to as \textit{data incubation} to grow the original collections and improve the generalization of recognizer training. 
This capability is enabled by a controllable generative model that is used to augment training with synthesized data covering underrepresented content and rare styles (\figref{figure:illustration}).

How does a controllable generative model improve data efficiency for a downstream handwriting recognizer?
After all, we learn the generative model with the same dataset that we use to train the recognizer. 
Our approach is based on the following observation:
The total number of handwriting samples that can be collected is limited and thus the diversity of contents and styles in the collected samples is also limited (\figref{figure:illustration}a).
By utilizing our controllable synthesizer that models the data distribution as a factored distribution of style and content, we can combine information across collected samples, complement the real training set with synthesized handwriting, and fill both the content and the style gaps in the collected training data (\figref{figure:illustration}b-d).

Effectively, we introduce prior knowledge about content and style being factorizable into the creation of synthetic training data. 
This inductive bias allows us to strategically increase the size of our training set to target a more balanced distribution of writing styles and content as well as training recognizers with better generalization.

Training with synthetic data, however, comes with its own challenges. 
Synthetic data might introduce a shift from the \textit{empirical distribution} that is formed by the collected real data. %
While this shift can be due to the artifacts produced by the generative model, it can also come from generating content and styles that are underrepresented or missing in the collected data, \ie, shifting the empirical distribution shown in \figref{figure:illustration}a to that in \figref{figure:illustration}d. 
In this aspect, this distribution shift can be seen both as a problem but also as a feature. 
It is critical to make careful and deliberate choices about which data to synthesize. 
If we only synthesize handwriting from well-formed English sentences with highly legible, printed style, a recognizer trained on this data will end up performing poorly on messy, cursive styles and content including slang, uncommon words, or abbreviations.
On the other hand, we can carefully choose which data to synthesize and with that we will be able to close the gaps in the empirical distribution.

In this paper, we demonstrate a systematic recipe to train a handwriting recognizer with data synthesized by a controllable generative model. 
This recipe enables us to examine and understand the problems associated with the synthetic data.
We build on the recently-proposed controllable generative model~\cite{chang2021style}, which separately controls the content and the handwriting style.
As we will demonstrate, utilizing data incubation significantly improves our multilingual handwriting recognizer, achieving a 66\% reduction in Character Error Rate when compared to a recognizer trained with only original data.

\section{Related work}

Online handwriting recognition is a field of high relevance in mobile computing because 
(a) an increasing number of devices are equipped with styluses making handwritten input a natural interface,
(b) many languages are harder to enter with a keyboard than Latin scripts because the size of their alphabet or the use of grapheme clusters makes building intuitive keyboards difficult. 
While early work in online handwriting recognition used segment-and-decode methods \cite{newton,pittman:2007,keysers2016multi,jaeger2001online}, recent work trends toward end-to-end approaches utilizing Convolutional Neural Networks (CNNs) and Recurrent Neural Networks (RNNs), coupled with Connectionist Temporal Classification (CTC) decoding~\cite{graves2006connectionist,plamondon2000online,carbune2020fast,graves-online-nips2008,frinken2015deep}.

Online handwriting synthesis models~\cite{chang2021style,zhang2018separating,bhunia2021handwriting,graves2013generating,aksan2018deepwriting,kotani2020generating} are sequence-to-sequence models that take input text (a sequence of characters) and output handwriting (a sequence of strokes).
To improve downstream recognizers, it is important for the synthesized handwriting to contain few artifacts and a wide range of handwriting styles --- from highly legible, printed-style handwriting to less legible, cursive handwriting.

Synthetic data augmentation for training machine learning models is not new in the literature. 
Most of the reference papers focus on image~\cite{xie2020adversarial,pouransari2021extracurricular} and speech~\cite{laptev2020you,li2018training,rosenberg2019speech,zheng2021using,mimura2018leveraging} synthesis, and obtain modest improvements. 
To the best of our knowledge, this paper is the first to use a controllable generative model to synthesize online handwriting and explicitly control the content, style, and variation of the generation process to significantly improve the accuracy of downstream handwriting recognizers.

\section{Generative and recognizer models}
\vspace{-0.2cm}
\label{sec: models}

\begin{figure}[t]
	\centering
	\includegraphics[width=0.9\linewidth]{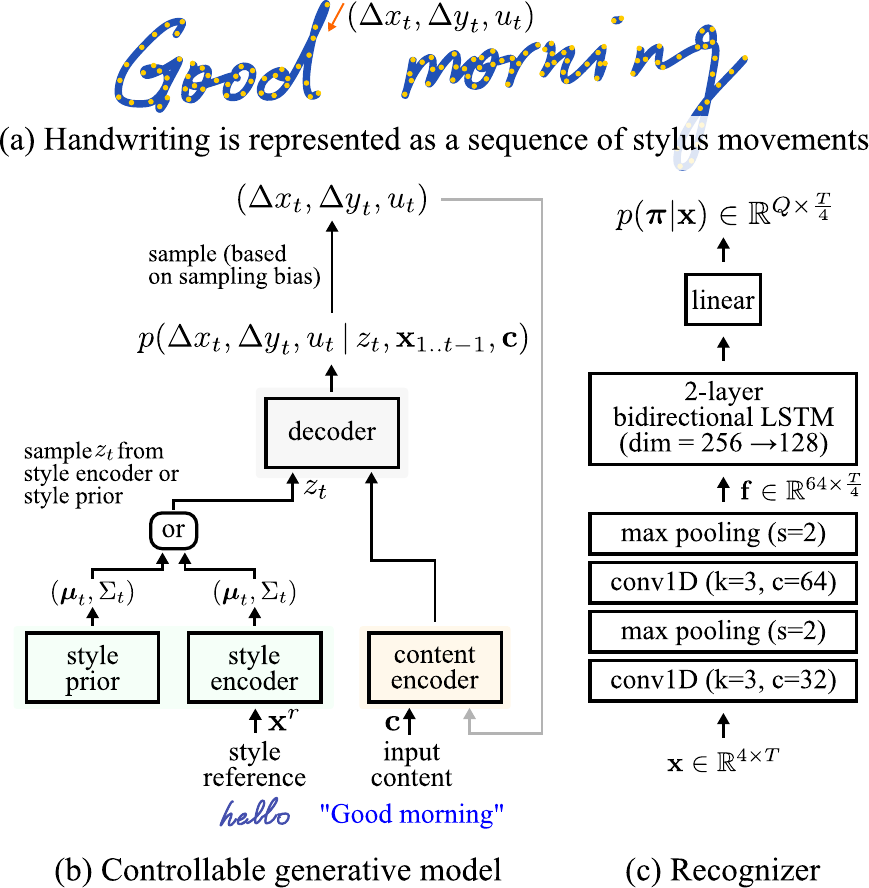}
	\vspace{-2.2mm}
	\caption{
	(a) Online handwriting is a sequence of stylus movements.
    (b) Our handwriting generative model takes an input content $\c$ and outputs handwriting $\x$.  The handwriting style is controlled using a latent style variable $z_t$. The model can either mimic the style of an existing style reference handwriting or it can sample a new style from a prior distribution to generate unseen styles. 
	(c) Our recognizer takes an input handwriting $\x$ and outputs the probabilities of all symbols at each time step, $p(\pi | \x)$, which is converted to $p(\c | \x)$ by CTC decoding.
	}
	\vspace{-2mm}
	\label{figure: arch}
\end{figure}

\subsection{Problem formulation and dataset}

An online handwriting sample $\x = \left[ (\dx_t, \dy_t, u_t) \right]_{t=1 \dots T}$ is a sequence of stylus movements on a writing surface, where $\dx_t \, {\in} \,  \R$ and $\dy_t \, {\in} \, \R$ are the movement of the stylys in $x$ and $y$ directions respectively at time $t$, and $u_t \, {\in} \, \{0, 1\}$ is a binary variable indicating whether the stylus touches the surface during the movement, as shown in \figref{figure: arch}a.
The corresponding content of the sample $\c = \left[ c_1, \dots c_M \right]$ is a sequence of characters, where $c_m$ is represented as a one-hot vector of dimension $Q$ (the total number of characters). 
Note that in general $\x$ and $\c$ have different lengths.  %

We use a subset of a proprietary dataset collected for research and development.  
The subset consists of 600k online handwriting samples written by 1,500 people.
Half of the samples (300k) is in English, and the rest is in French, German, Italian, Spanish, and Portuguese.

\subsection{Controllable generative model }

We use the controllable generative model proposed in \cite{chang2021style}.
Given $N$ real training samples, $\cX = \left\{  (\x^i, \c^i) \right\}_{i=1 \dots N}$, the generative model learns a distribution of handwriting samples, $p(\x | \c, \z)$, conditioned on the content $\c$ and the style $\z$.
In other words, given a target content $\c$ and a style $\z$, we can generate new handwriting by sampling from the distribution.
Note that the style $\z$ is modeled as a latent variable with a (learnable) prior distribution. 
Fitting a prior distribution for style is important, because it enables the model to create new styles (\ie, $\z$ not associated with any training sample).

\figref{figure: arch}b shows an overview of the model architecture. 
We model in an auto-regressive manner the output distribution $p(\dx_t, \dy_t, u_t | z_t, \x_\past, \c)$ as $p(\dx_t, \dy_t | \cdot) p(u_t | \cdot)$, where $p(x_t, y_t | \cdot)$ is a mixture of 20 bivariate Gaussian distributions modeling the movement of the stylus, and $p(u_t |  \cdot)$ is a Bernoulli distribution modeling whether the stylys touches the surface.
The prior distribution of $z_t$, $p(z_t | \x_\past, \c)$, is a multivariate Gaussian distribution $\mN(z_t | \bmu_t, \Sigma_t))$, where $\Sigma_t$ is a diagonal covariance, and $\bmu_t$ and $\Sigma_t$ are functions of $\x_\past$ and $\c$ (generated by a linear layer).
We refer to \cite{chang2021style} for detailed description of the model.
We trained two generative models: an English model using only the English training data and a multilingual model using training data from all six languages.

One important property of the generative model is that we can control the variation of the synthesized data by scaling the standard deviations of the output distribution $p(\dx_t, \dy_t, u_t | z_t, \x_\past, \c)$.
Intuitively, using a small standard deviation generates handwriting close to the mean of the distribution, which results in neat (average-style) handwriting; using a large standard deviation results in more variations in the synthesized handwriting.
When synthesizing, we use \textit{sampling bias} $b \ge 0$ to scale individual standard deviations and the categorical distribution in the mixture of the Gaussian distributions, following the rule in~\cite{graves2013generating}. 
A small sampling bias produces large variations in handwriting, and a large sampling bias results in small variations and neat/typical handwriting.

\vspace{-0.3cm}

\subsection{Handwriting recognizer}
\vspace{-0.1cm}

A handwriting recognizer learns a distribution of the content, $p(\c | \x)$, given an input handwriting sequence $\x$.
As shown in \figref{figure: arch}c, our handwriting recognizer is a 2-layer convolution network followed by a 2-layer bidirectional LSTM and a linear layer.
The input to the recognizer is a sequence containing $u_t, \sin(\theta_t), \cos(\theta_t), \ell_t$ for $t = 1, \dots, T$, where $\theta$ and $\ell$ are the angle and length of the stylus movement at time $t$, \ie, they are features computed from $\dx_t$ and $\dy_t$.
The network outputs the probabilities of each character (including a \textit{blank} symbol) at every time $t$.
We use CTC loss~\cite{graves2006connectionist} to train the model.

\section{Synthesizing training data}
\label{sec: cases}

If data is not synthesized carefully, the domain gap between real and synthetic data can deteriorate the performance of the recognizer on real data. Here, we present a systematic framework to identify problems with synthetic data and optimize  parameters of the generative model to minimize the gap. Our recipe uses a similar procedure as \cite{shmelkov2018good}  to evaluate the generation quality of a generative adversarial network.

Given a synthetic dataset (randomly split into training, validation, and test sets), we train three recognizers: one using only the real training data, one using only the synthetic training data, and one using both.
Each model is evaluated using Character Error Rate (CER) on both the real and synthetic test sets. We use the notation \WERtestXY to indicate the CER for a model trained on training set $\text{t}$ and evaluated on set $\text{v}$, where $\text{t},\text{v} \in \{\text{r}, \text{s}, \text{b}\}$ indicating \textbf{r}eal, \textbf{s}ynthetic, and \textbf{b}oth data sets, respectively.

Note that the real training set (empirical distribution) is only a finite sample approximation to the true data distribution and might be biased due to the data collection procedure or lack certain content, style, or other variations. Here, we refer to the empirical distribution as $\mS_r$, synthetic data distribution as $\mS_s$ and the true data distribution as $\mS_t$. We assume that the real test set is unbiased. The following are the \emph{common} cases:

\paragraph{Case 1: $\mS_s \not \subseteq \mS_t$.} Synthetic data contains artifacts or variations that do not exist in the true distribution. These artifacts are equivalent to introducing noisy samples to the training set. In an extreme case, the artifacts may correlate with the content; therefore, the recognizer trained on the synthetic data can exploit these artifacts and will not generalize. The extreme case can be diagnosed by lower \WERtestSS, but high \WERtestSR and \WERtestRS. 

\paragraph{Case 2:  $\mS_s \subseteq \mS_t \ \ \& \ \ \mS_s \subset \mS_r$.} Synthetic data is a subset of the true distribution, but fails to capture all the variations that exist in the real dataset (empirical distribution). The model trained on synthetic data will fail to generalize well. For example, when both real and synthetic data share the same corpus but the synthetic data only contain printed-style handwriting, the recognizer trained on synthetic data will perform poorly on the real data. Similarly, as we have discussed in \secref{sec: models}, when we use a high sampling bias to generate data, the synthetic dataset contains mostly neat and typical handwriting. Thus, the recognizer will perform poorly on real handwriting, which can be difficult to read sometimes. This case can be diagnosed by lower \WERtestSS and \WERtestRS, but high \WERtestSR.

\paragraph{Case 3:  $\mS_s \subseteq \mS_t \ \ \& \ \ \mS_s \not \subseteq \mS_r$.}  
Synthetic data is a subset of the true distribution and contains variations (\eg missing style and content) that are not covered in the real dataset. This is the scenario where the recognizer will  benefit  from  the  synthetic  data. By filling in missing data, synthetic data improves the performance of the recognizer leading to lower \WERtestSR (and \WERtestSS) than \WERtestRR and \WERtestRS.

To mitigate the problems discussed in case 1 and case 2, either the generative model or the sampling process has to be improved accordingly. When the generative model exactly models the true distribution, \WERtestSR and \WERtestSS should be identical. In practice, this is difficult to achieve; however, choosing the generative model such that  \WERtestSR and \WERtestSS are close is a good strategy to minimize the gap between synthetic and real data. Moreover, in general, neither
case 2 nor case 3 are exact (synthetic and real data contains different variations of the true distribution); therefore, combining real data with synthetic data (\WERtestBR) improves over using synthetic data (\WERtestSR) or real data (\WERtestRR) alone. In the experiments section, we analyze these cases based on the sampling bias of the generative model.

\section{Experiments}
\label{sec: experiment}

\begin{figure}
	\centering
	\begin{subfigure}{.48\linewidth}
		\centering
		\includegraphics[width=\linewidth]{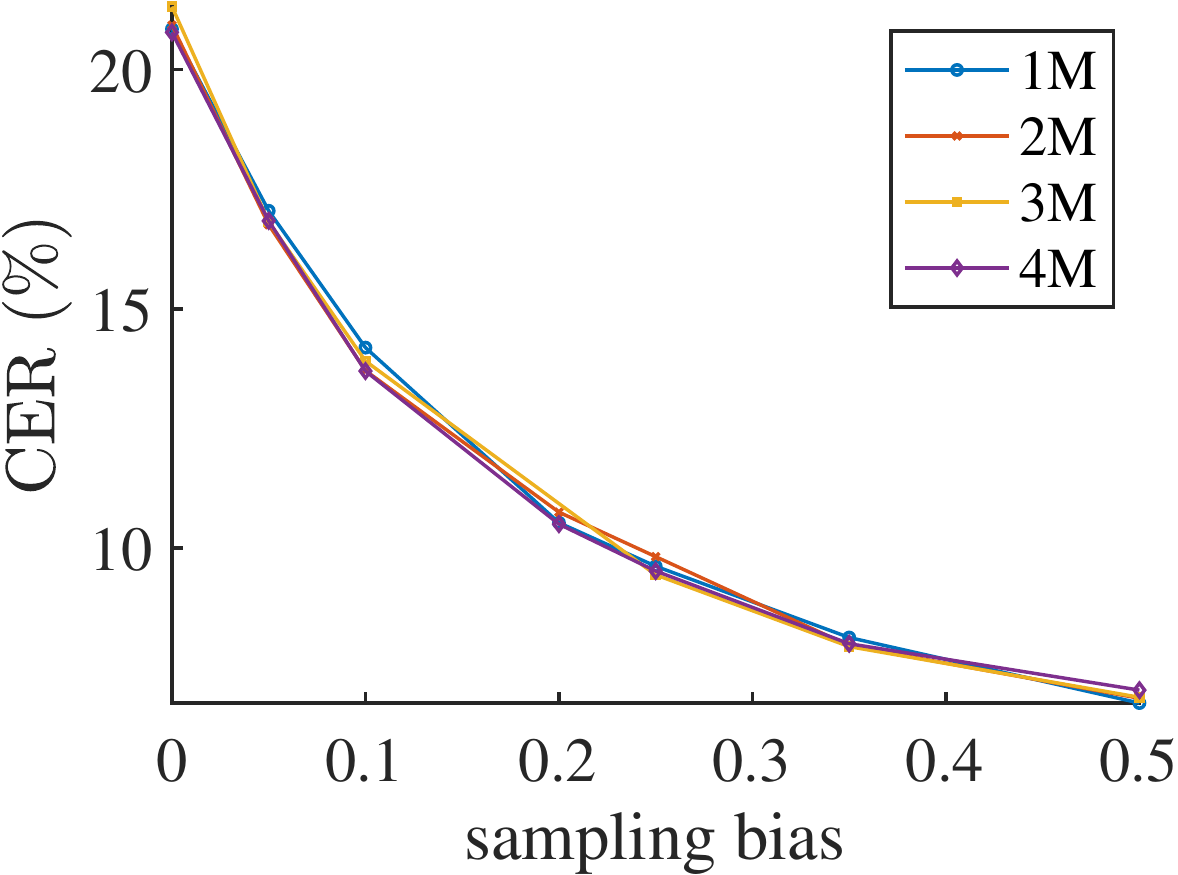}
		\caption{\WERtestRS on English}
		\label{figure: exp wer test rs}
	\end{subfigure}
	\hfill
	\begin{subfigure}{.48\linewidth}
		\centering
		\includegraphics[width=\linewidth]{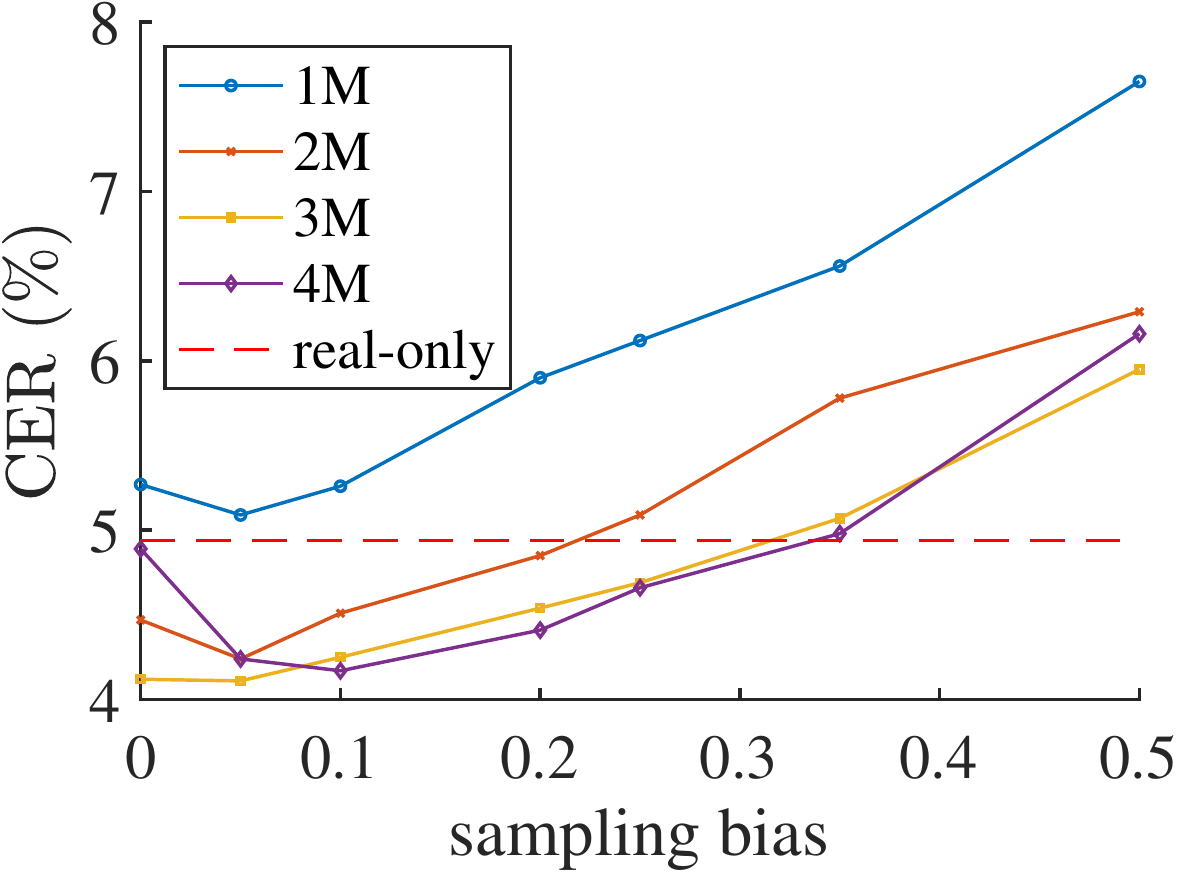}
		\caption{\WERtestSR on English}
		\label{figure: exp wer test sr}
	\end{subfigure}
	\\
	\vspace{2mm}
	\begin{subfigure}{.48\linewidth}
		\centering
		\includegraphics[width=\linewidth]{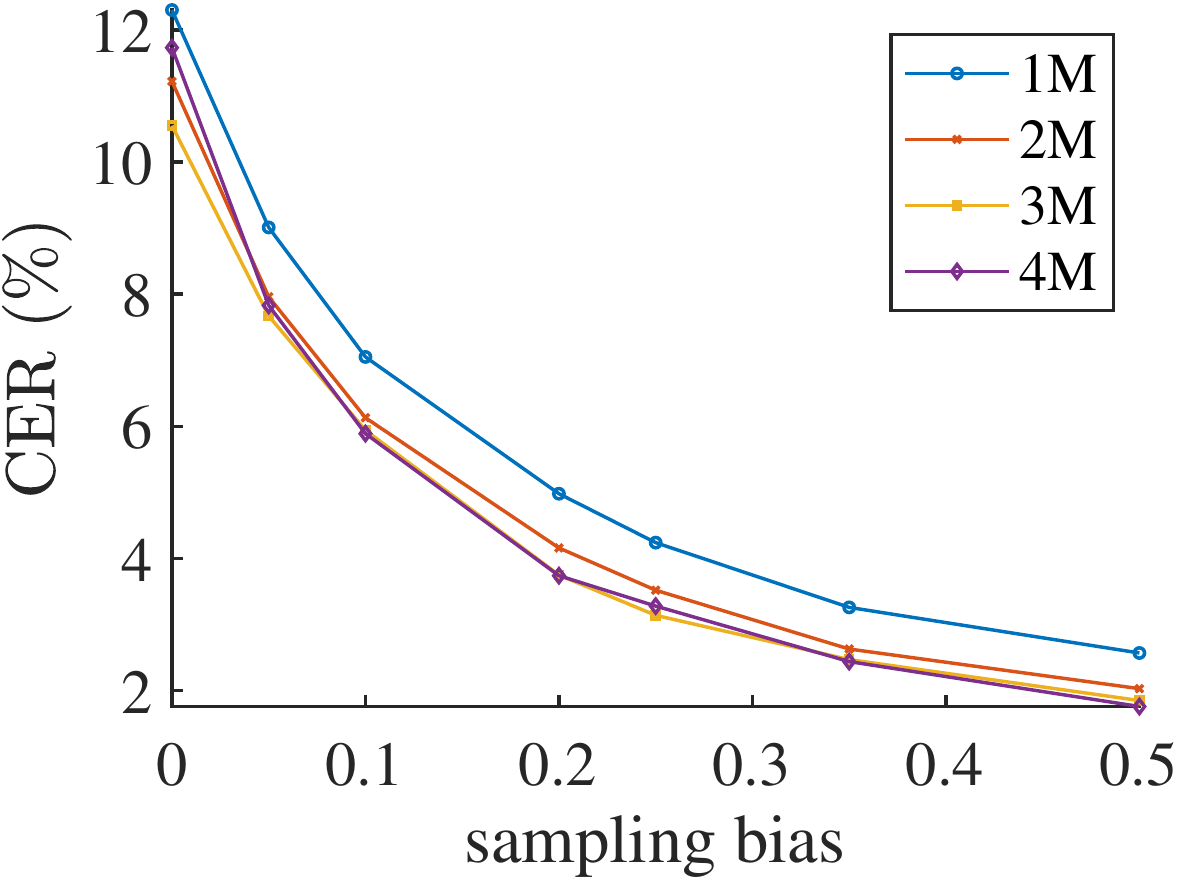}
		\caption{\WERtestSS on English}
		\label{figure: exp wer test ss}
	\end{subfigure}
	\hfill
	\begin{subfigure}{.48\linewidth}
		\centering
		\includegraphics[width=\linewidth]{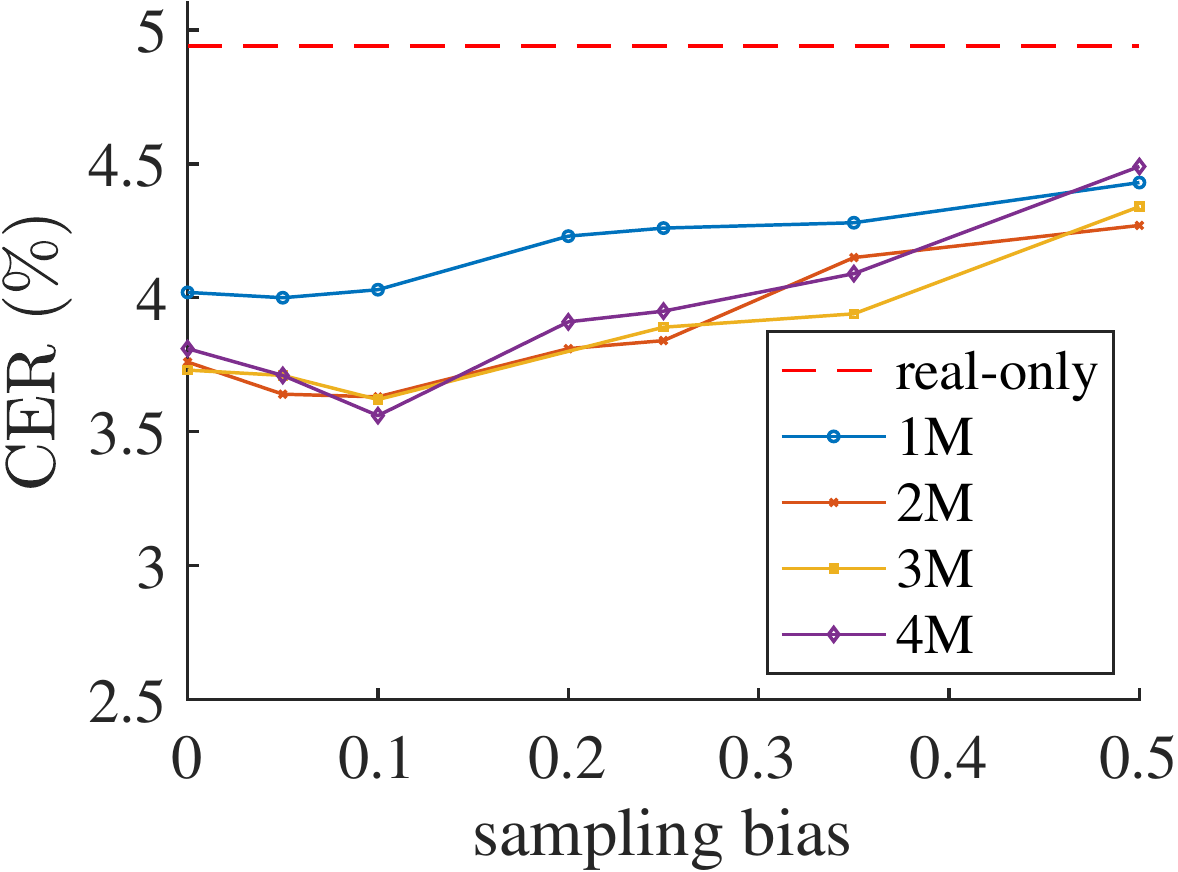}
		\caption{\WERtestBR on English}
		\label{figure: exp wer test br}
	\end{subfigure}
	\caption{Effect of sampling bias and synthetic dataset size on English recognizers.}
	\label{figure: exp cer}
\end{figure}

We first use our proposed framework to evaluate the effect of style diversity.
As discussed, increasing the sampling bias reduces the style diversity of synthetic data.
\figref{figure: exp cer} shows the CERs of the English recognizer when varying the sampling bias and the amount of synthetic training data.
To avoid introducing new content, the  data was synthesized using the text corpus of the original training data.
The results indicate that:
(1) A large sampling bias (bias $\ge$ 0.4; \ie neat training data) leads to low \WERtestRS and high \WERtestSR, indicating a decrease in the diversity of the synthetic data (\figref{figure: exp wer test rs}), corresponding to Case 2.
(2) A small sampling bias (0.05 $\le$ bias < 0.4) increases the diversity of the data, covering a large distribution of styles, leads to lower  \WERtestSR and \WERtestBR than \WERtestRR, corresponding to Case 3.
Due to the increase in data diversity, we also observe an increase in \WERtestRS and \WERtestSS.
(3) Further reducing the sampling bias (bias < 0.05) leads to an increase in \WERtestSR and \WERtestBR due to the occasional generation of out-of-distribution samples, corresponding to Case 1. %
When we use a very small sampling bias, we sample from a mixture of Gaussian distributions with large standard deviations.
Therefore, the larger standard deviations make it more likely to generate out-of-distribution samples.

These results demonstrate the effectiveness of data incubation to improve recognizers and of the proposed recipe to analyze synthetic data and optimize synthesis.
By selecting the right sampling bias, we are able to reduce the CER of the English recognizer by 28\% compared to the model trained only using real data (Table~\ref{table: final}) because we are able to increase the diversity of writing styles.

Further, for our multilingual recognizer, we show that using additional content helps improve the CER even further in a scenario where we don't have good coverage of all relevant content.
When training our system with the same text corpus, we obtain an improvement of 46\% on the normal multilingual test corpus.
However, in this scenario, we only have 60k training samples for each of the non-English languages (compared to 300k for English). Therefore we increase the content diversity by synthesizing from a large multilingual text corpus. This leads to an improvement of 66\% (Table~\ref{table: final}).

Finally, we also show results on a special patterns (SP) evaluation set containing content like URLs, emails, addresses, and hashtags which are underrepresented in the collected training data. 
As shown in the last column of Table~\ref{table: final}, expanding the synthesis corpus enables our multilingual model to also achieve 66\% reduction in CER compared to the model trained using only the real data.

\begin{table}[t]
	\caption{
		CERs on real test data for the English and multi-language setup. We report which training data was used, the size of the training data, the CER, and relative improvement. For the multi-language setup, we also report the results on a special patterns (SP) dataset containing URLs, made-up addresses, and made-up   email addresses.
	}
	\label{table: final}
	\small
	\centering
	\def\rowpad{2mm} 
	\setlength{\tabcolsep}{3.6pt}
	\vspace{-2mm}
	\begin{adjustbox}{max width=\linewidth}
	
	\small
	\begin{tabular}{llcclrr}
	\toprule
	      &         &             & \multicolumn{2}{c}{Test} & \multicolumn{2}{c}{SP Test} \\
	\cmidrule(lr){4-5} \cmidrule(lr){6-7}       
	Setup & tr.\ data & $|$tr.\ data$|$ & CER & impr. & CER & impr.\\
	\midrule
	English & real       &  0.3 & 4.9 &      \makecell{-}  &  n/a & n/a         \\
	        & synthetic  &  4.0 & 4.2 & $\downarrow$ 16\% & n/a & n/a \\
	        & both       &  4.3 & \bf 3.6 &\bf $\downarrow$ 28\% & n/a & n/a \\
	\midrule
	Multi-Lang &   real &0.6 & 8.2 & \makecell{-} & 18.5 & \makecell{-} \\
	 & synthetic&3.0 & 5.2 &$\downarrow$ 37\% & 18.9  &$\uparrow$ 2\%\\
	&     both&  3.6&4.5 &$\downarrow$ 46\% & 16.9 &$\downarrow$ 9\%\\
	&     both (ext)&  8.1& \bf 2.8 &\bf $\downarrow$ 66\% & \bf 6.3 &\bf $\downarrow$ 66\%\\
	\bottomrule
	
	\end{tabular}
	\end{adjustbox}
\end{table}

\section{Conclusion}
We presented data incubation{\textemdash}a new framework to improve recognition by (i) training a controllable generative model to synthesize missing data, (ii) optimizing data synthesis, and (iii) training recognition models by strategically combining synthetic and real data. We applied this framework to handwriting recognition and achieved a $66\%$ reduction in CER.

\bibliographystyle{IEEEbib}
\bibliography{main}

\end{document}